\documentclass[11pt]{article}
\usepackage[utf8]{inputenc}
\usepackage[T1]{fontenc}
\usepackage{amsmath,amssymb}
\usepackage{graphicx}
\usepackage{booktabs}
\usepackage{hyperref}
\usepackage[margin=1in]{geometry}
\usepackage{natbib}
\usepackage{xcolor}
\usepackage{enumitem}

\hypersetup{colorlinks=true,linkcolor=blue,citecolor=blue,urlcolor=blue}

\title{Verbal Confidence Saturation in 3--9B Open-Weight Instruction-Tuned LLMs:\\A Pre-Registered Psychometric Validity Screen}

\author{Jon-Paul Cacioli\\
Independent Researcher, Melbourne, Australia\\
ORCID: 0009-0000-7054-2014\\
\texttt{https://github.com/synthiumjp/koriat}}

\date{April 2026}

\begin{document}
\maketitle

\begin{abstract}
Verbal confidence elicitation is widely used to extract uncertainty estimates from LLMs. We tested whether seven instruction-tuned open-weight models (3--9B parameters, four families) produce verbalised confidence that meets minimal validity criteria for item-level Type-2 discrimination under minimal numeric elicitation with greedy decoding. In a pre-registered study (OSF: \url{https://osf.io/azbvx}), 524 TriviaQA items were administered under numeric (0--100) and categorical (10-class) elicitation to eight models at Q5\_K\_M quantisation on consumer hardware, yielding 8,384 deterministic trials. A psychometric validity screen (Appendix~A) was applied to each model--format cell. All seven instruct models were classified Invalid on numeric confidence (H2 confirmed, 7/7 vs.\ predicted $\geq$ 4/7), with a mean ceiling rate of 91.7\% (H1 confirmed). Categorical elicitation did not rescue validity. Instead, it disrupted task performance in six of seven models, producing accuracy below 5\% (H4 not confirmed). Token-level logprobability did not usefully predict verbalised confidence under the observed variance regime (H5 confirmed, mean cross-validated $R^2 < 0.01$). Within the reasoning-distilled model, reasoning-trace length showed a strong negative partial correlation with confidence ($\rho = -.36$, $p < .001$), consistent with the Reasoning Contamination Effect. These results do not imply that internal uncertainty representations are absent. They show that minimal verbal elicitation fails to preserve internal signals at the output interface in this model-size regime. Psychometric screening should precede any downstream use of such signals.
\end{abstract}

\section{Introduction}

Verbal confidence elicitation is widely used to extract uncertainty estimates from large language models \citep{xiong2023,tian2023,steyvers2025}. The premise is that the verbalised number reflects an internal signal that discriminates correct from incorrect responses. Recent mechanistic work supports this in part. \citet{kumaran2026} showed that verbal confidence in Gemma 3 27B is computed via cached retrieval at answer-adjacent positions and contains variance not explained by token log-probabilities.

The practical utility of verbal confidence depends on whether the elicited signal carries item-level information in the deployment context. Three concurrent lines of evidence establish a readout problem. \citet{miao2026} showed that calibration and verbalised confidence are encoded in orthogonal directions in the residual stream. \citet{wang2026} documented confidence saturation on TriviaQA and SimpleQA. \citet{seo2026} identified answer-dependent confidence generation as a driver of overconfidence. \citet{yang2024} found that verbalised confidence in small open-weight models can be near-independent from accuracy, though this depends strongly on the elicitation method.

Most calibration work assumes the confidence distribution spans the scale and varies with correctness \citep{geng2024,steyvers2025b}. This study tests that assumption directly. We apply a psychometric validity screen to classify model--format combinations as Invalid, Indeterminate, or Valid before any calibration is attempted. The question is not how miscalibrated the signal is, but whether it meets minimal criteria for item-level Type-2 use.

We treat saturation as a validity failure rather than a calibration failure. A distribution collapsed to the ceiling cannot support item-level discrimination because the ordinal relationships between trials have been lost at elicitation. No post-hoc rescaling can recover what was never emitted.

This study evaluates default interface behaviour under minimal elicitation with greedy decoding. It does not test the capacity of these models to express uncertainty under structured or scaffolded prompts. A model that fails the screen here may produce a valid signal under a richer elicitation regime. That is a separate empirical question. Validity is a property of the model--probe--task interaction \citep{cacioli2026e,rust2021}, not an intrinsic property of the model.

\subsection{Hypotheses}

Four confirmatory hypotheses were pre-registered (OSF: \url{https://osf.io/azbvx}).

\textbf{H1 (saturation prevalence).} The mean proportion of trials with numeric confidence $\geq 95\%$ across seven instruct models exceeds 60\%.

\textbf{H2 (validity screening).} At least four of seven instruct models are classified Invalid on numeric confidence under the validity protocol (Appendix~A).

\textbf{H4 (format rescue).} Among models classified Invalid under numeric elicitation, at least two are reclassified as non-Invalid under categorical elicitation.

\textbf{H5 (logprob--confidence independence).} Mean cross-validated $R^2$ of ridge regression predicting verbal confidence from length-normalised logprobability is below 0.20 in both conditions, indicating no usable predictive relationship.

H3 was retired pre-registration after sanity data showed the base model does not produce verbalised confidence under continuation prompting (see pre-registration \S10).

\section{Methods}

\subsection{Models}

Eight open-weight LLMs were evaluated, all run as Q5\_K\_M GGUF quantisations via llama-cpp-python 0.3.16 with Vulkan backend on an AMD RX 7900 GRE 16\,GB (Table~\ref{tab:models}).

\begin{table}[ht]
\centering
\caption{Model sample.}
\label{tab:models}
\begin{tabular}{llllr}
\toprule
ID & Model & Family & Params & Role \\
\midrule
M1 & Meta-Llama-3-8B & Llama 3 base & 8B & Exploratory \\
M2 & Meta-Llama-3-8B-Instruct & Llama 3 instruct & 8B & Confirmatory \\
M3 & Meta-Llama-3.1-8B-Instruct & Llama 3.1 instruct & 8B & Confirmatory \\
M4 & Mistral-7B-Instruct-v0.3 & Mistral & 7B & Confirmatory \\
M5 & Qwen2.5-3B-Instruct & Qwen 2.5 & 3B & Confirmatory \\
M6 & Qwen2.5-7B-Instruct & Qwen 2.5 & 7B & Confirmatory \\
M7 & Gemma-2-9b-it & Gemma 2 & 9B & Confirmatory \\
M8 & DeepSeek-R1-Distill-Llama-8B & DeepSeek R1 & 8B & Confirmatory \\
\bottomrule
\end{tabular}
\end{table}

M1 is a base model retained for one exploratory analysis (E-base) comparing logprob distributions between the base and instruct versions of Llama-3-8B. M1 does not contribute to any confirmatory hypothesis.

\subsection{Substrate and design}

524 items from TriviaQA rc.nocontext validation split \citep{joshi2017}, drawn deterministically with \texttt{numpy.random.default\_rng(seed=42)}. Design: 8 models $\times$ 2 conditions (NUM, CAT) $\times$ 524 items = 8,384 trials. Confirmatory sample: 7 instruct models $\times$ 2 $\times$ 524 = 7,336 trials. Inference was greedy (temperature = 0), deterministic (seed = 42), with logprobs collected for the top 5 tokens at each position.

Greedy decoding isolates the model's maximum-likelihood output. Sampling introduces stochastic variance that conflates confidence-expression behaviour with sampling noise. Many real deployments do not use pure greedy decoding. The findings characterise default interface behaviour, not model capacity.

\subsection{Elicitation conditions}

\textbf{Numeric (NUM).} System prompt: ``You are answering trivia questions. After your answer, state your confidence as a percentage from 0 to 100.''

\textbf{Categorical (CAT).} System prompt: ``Classify your confidence into one of the following classes based on how likely the answer above is to be correct (NO REASONING OR EXPLANATION):'' followed by a 10-class ordinal scale from ``No chance'' (0.0--0.1) to ``Almost certain'' (0.9--1.0).

M1 used a continuation prompt (Q: \{question\} / A:) with no confidence elicitation.

\subsection{Validity screening protocol}

Each model--condition cell was screened using a psychometric validity protocol adapted from clinical assessment practice (MMPI-3: \citealt{benporath2020}; PAI: \citealt{morey1991}). The full protocol is provided in Appendix~A. Criterion validity was demonstrated in \citet{cacioli2026f}, where Invalid-classified models showed mean AUROC$_2$ = .357 versus .624 for Valid-classified models ($d = 2.81$).

Continuous confidence values are binarised at 0.50, producing a 2$\times$2 contingency table. Three indices are computed: $L = P(\text{high confidence} \mid \text{incorrect})$, $F_p = P(\text{low confidence} \mid \text{correct})$, and $\text{RBS} = F_p - (1 - L)$. A degeneracy pre-check classifies any cell with $> 95\%$ of binarised responses in a single category as Invalid without further analysis.

\subsection{Analysis plan}

All analyses were pre-registered (OSF: \url{https://osf.io/azbvx}, locked 15 April 2026). Cells with $> 30\%$ parse failure are excluded from confirmatory analyses. Parse failure is treated as missing at random. The MAR assumption is evaluated in E8 and the H1 sensitivity analysis (A8) tests robustness directly.

\subsection{Pre-registration deviations}

One deviation occurred. The original ParquetWriter class inferred its schema from the first flushed batch. When batches contained entirely null values in nullable columns, pyarrow inferred mismatched types. The fix replaced inferred schema with an explicit pyarrow schema. No collection logic, seeds, prompts, parsing, or records were affected.

\section{Results}

\subsection{Confirmatory hypotheses}

\begin{figure}[ht]
\centering
\includegraphics[width=0.95\textwidth]{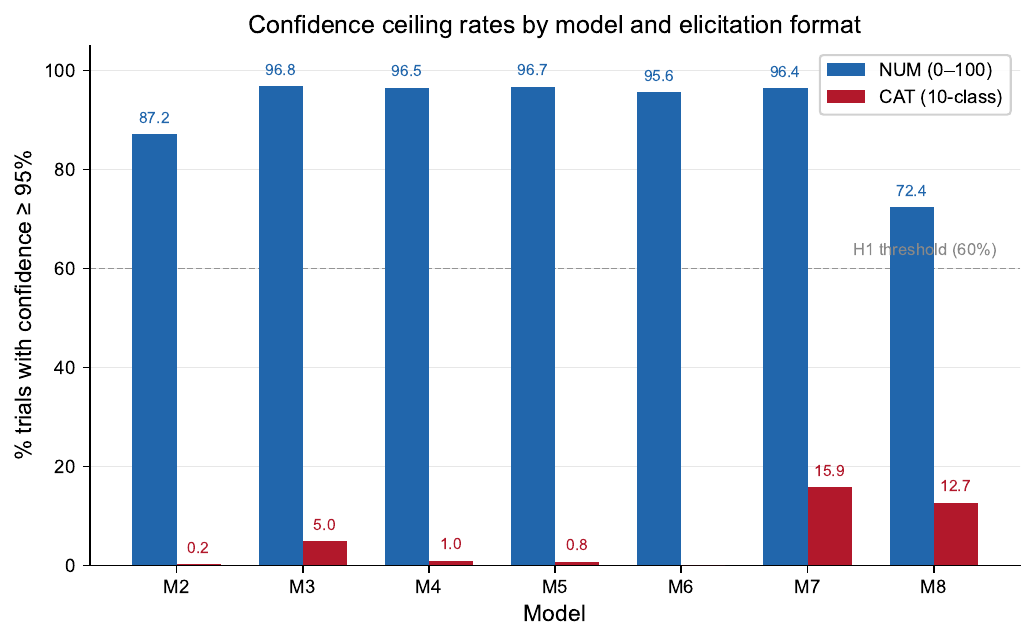}
\caption{Confidence ceiling rates by model and elicitation format. The dashed line shows the H1 threshold (60\%). All seven instruct models exceed 72\% on NUM. CAT ceiling rates are near zero for most models.}
\label{fig:ceiling}
\end{figure}

\textbf{H1: Saturation prevalence, confirmed.} Across the seven instruct models on numeric elicitation, the mean proportion of parse-success trials with confidence $\geq 95\%$ was 91.7\% (range 72.4--96.8\%). All seven exceeded the 60\% threshold individually (Figure~\ref{fig:ceiling}). The sensitivity analysis (A8), coding parse failures as non-variable responses, yielded 92.7\%.

\textbf{H2: Validity screening, confirmed.} All seven instruct models were classified Invalid on numeric confidence, exceeding the pre-registered threshold of $\geq 4$. Five models were classified Invalid via the degeneracy criterion ($> 95\%$ of binarised responses in a single category) with $L$ values ranging from 0.949 to 0.986. Two models (M2, M7) showed complete degeneracy: TRIN = 1.0, $L = 1.0$, zero low-confidence observations across 500+ trials.

\textbf{H4: Format rescue, not confirmed.} Zero models were reclassified from Invalid (NUM) to non-Invalid (CAT). Categorical elicitation introduced a different failure mode: task-performance collapse (Table~\ref{tab:cat}).

\begin{table}[ht]
\centering
\caption{Accuracy and parse rates under categorical elicitation.}
\label{tab:cat}
\begin{tabular}{lrrrrl}
\toprule
Model & $n$ success & $n$ correct & Accuracy & Parse rate & Validity tier \\
\midrule
M2 & 504 & 10 & 2.0\% & 96.2\% & Insufficient \\
M3 & 161 & 105 & 65.2\% & 30.7\% & Insufficient (\S6.6) \\
M4 & 523 & 13 & 2.5\% & 99.8\% & Insufficient \\
M5 & 503 & 21 & 4.2\% & 96.0\% & Insufficient \\
M6 & 521 & 4 & 0.8\% & 99.4\% & Insufficient \\
M7 & 523 & 1 & 0.2\% & 99.8\% & Invalid (degeneracy) \\
M8 & 71 & 18 & 25.4\% & 13.5\% & Insufficient \\
\bottomrule
\end{tabular}
\end{table}

The specific 10-class categorical prompt disrupted question-answering behaviour. Models that achieved 59--76\% accuracy under NUM dropped to 0.2--4.2\% under CAT. Whether simpler categorical formats would avoid this interaction failure is an open question \citep{yang2024}.

\textbf{H5: Logprob--confidence independence, confirmed.} Mean cross-validated $R^2$ was $-0.60$ on NUM and 0.03 on CAT, both well below the 0.20 threshold. The negative NUM mean is driven by M5 ($R^2_{\text{CV}} = -4.61$), where near-zero confidence variance produces degenerate folds. The median $R^2_{\text{CV}}$ on NUM was 0.06. Token-level logprobability does not usefully predict verbalised confidence under the observed variance regime.

\subsection{Type-2 AUROC$_2$}

\begin{figure}[ht]
\centering
\includegraphics[width=0.85\textwidth]{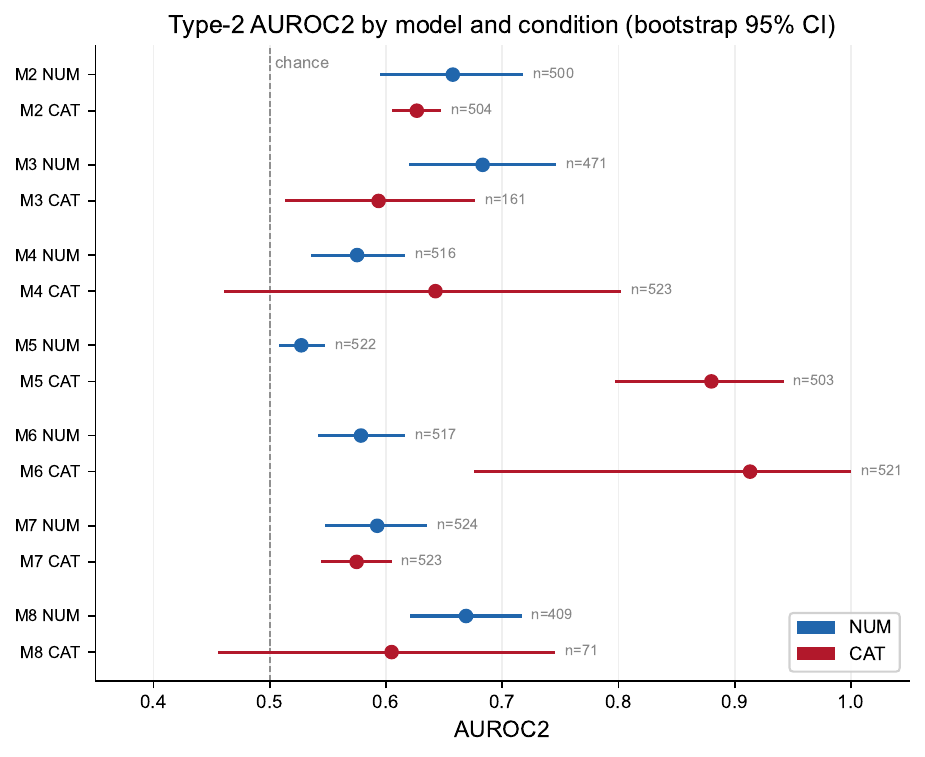}
\caption{Type-2 AUROC$_2$ by model and condition (bootstrap 95\% CI). NUM values are modestly above chance with tight CIs. CAT values for M4, M5, M6, M8 show wide CIs reflecting unreliable estimates from extreme base-rate imbalance.}
\label{fig:auroc}
\end{figure}

Despite universal validity failure on NUM, AUROC$_2$ values were modestly above chance (Figure~\ref{fig:auroc}), ranging from 0.527 (M5) to 0.683 (M3). This does not contradict the validity classification. AUROC$_2$ is a ranking metric driven by 3--13\% residual non-ceiling responses. A deployment system cannot act on a signal identical for 97\% of inputs.

\subsection{Exploratory analyses}

\begin{figure}[ht]
\centering
\includegraphics[width=0.85\textwidth]{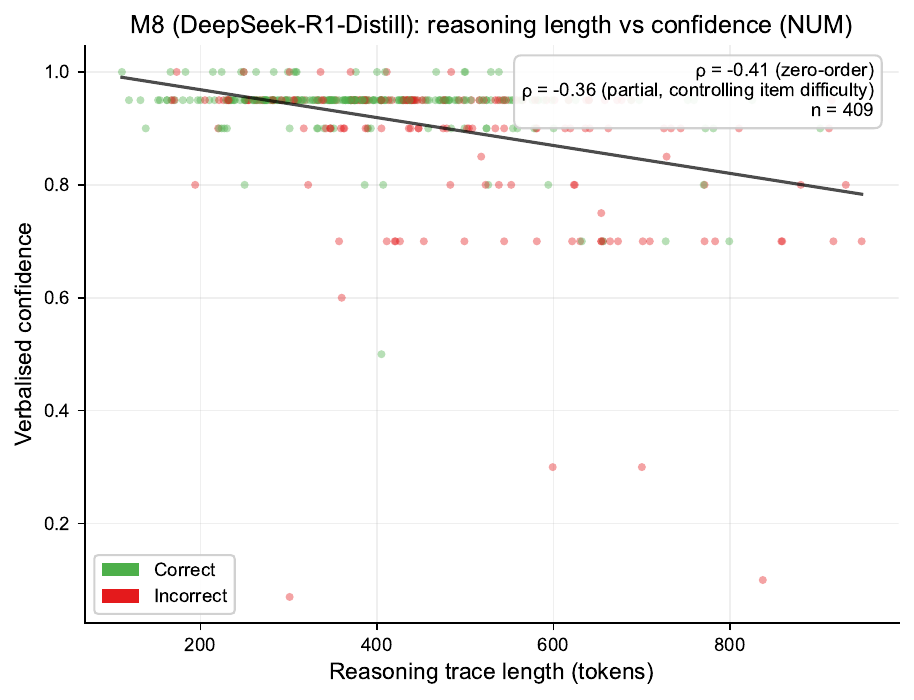}
\caption{M8 (DeepSeek-R1-Distill): reasoning-trace length vs.\ verbalised confidence (NUM). Points coloured by correctness. The negative slope persists after controlling for item difficulty ($\rho_{\text{partial}} = -.36$).}
\label{fig:e5}
\end{figure}

\textbf{E5: Reasoning contamination probe.} Within M8 NUM ($n = 409$), reasoning-trace length showed a negative zero-order Spearman correlation with verbalised confidence ($\rho = -.41$, $p < .001$). After controlling for item difficulty, the partial correlation remained strong ($\rho = -.36$, $p < .001$; Figure~\ref{fig:e5}). This is consistent with \citeauthor{miao2026}'s (\citeyear{miao2026}) Reasoning Contamination Effect and with a simpler account where difficult items require more tokens and receive lower confidence.

\textbf{E3: Item-level sensitivity.} Item difficulty correlated with mean confidence across instruct models at $\rho = .50$ ($p < .001$, $n = 524$). The models carry item-level information. It does not survive the verbalised readout under minimal elicitation.

\textbf{E8: MAR plausibility check.} Parse failure was higher for correct trials (18.8\%) than incorrect (10.1\%), indicating the MAR assumption is not strictly met. The A8 sensitivity analysis is consistent with the saturation pattern being conservative (mean ceiling rate increased from 91.7\% to 92.7\%). Non-random missingness remains a limitation.

\textbf{E9: Split-half stability.} Validity classifications agreed across random half-splits on 14/14 instruct model--condition cells. 100\% agreement at 262 items per split.

\section{Discussion}

\subsection{Summary}

Under minimal numeric elicitation with greedy decoding, verbal confidence from seven instruction-tuned open-weight models in the 3--9B range fails to meet minimal validity criteria. Every model was classified Invalid on numeric confidence. A 10-class categorical prompt did not rescue validity. It disrupted task performance. Token-level logprobability did not usefully predict the verbalised signal. These findings hold across four model families and a reasoning-distilled model.

\subsection{Saturation as validity failure}

The descriptive pattern is stark before any protocol is invoked. Six of seven instruct models assigned confidence $\geq 95\%$ on more than 87\% of parse-success trials. Two models (M2, M7) produced zero low-confidence responses across 500+ trials. The confidence output was a constant. The validity protocol (Appendix~A) formalises what the raw distributions already show.

This is a validity failure, not a calibration failure. The qualification is important: this applies to the elicited readout under these conditions, not to internal representations. E3 confirms that these models carry item-level information correlated with confidence ($\rho = .50$). The information exists. It does not survive the verbalised readout under minimal elicitation.

The saturation observed here is not an intrinsic property of 3--9B models. It is a property of the model--probe--task interaction under these conditions. Whether richer elicitation can unlock a valid signal is a separate question.

\subsection{Two failure modes}

The H4 result reveals two distinct failure modes. \textbf{Saturation failure (NUM):} the model produces answers and confidence ratings, but confidence is compressed to the ceiling. $L \geq 0.95$ in most models. \textbf{Interaction failure (CAT):} the specific 10-class categorical prompt disrupted question-answering behaviour. Models that achieve 59--76\% accuracy on NUM dropped to 0.2--4.2\% on CAT. This is not evidence that categorical elicitation is inherently flawed. M3 maintained 65.2\% accuracy under the same prompt. This particular prompt is poorly tolerated by most models in this size range.

\subsection{AUROC$_2$ and validity}

AUROC$_2$ values on NUM range from 0.527 to 0.683. This does not contradict the Invalid classification. When 87--97\% of trials sit at the ceiling, ranking is driven by 3--13\% residual responses. The validity indices decompose what AUROC$_2$ aggregates over.

\subsection{Deployment implications}

Any system using verbal confidence from small open-weight instruct models for abstention, routing, or safety decisions under minimal elicitation is building on a degenerate signal. Verbal confidence signals should be screened for validity before calibration or selective prediction metrics are computed. The protocol (Appendix~A) requires only the raw confidence outputs and correctness labels already available in any benchmark evaluation.

\subsection{Limitations}

\textbf{Elicitation regime.} One minimal numeric prompt and one 10-class categorical prompt under greedy decoding. Many real deployments do not use pure greedy decoding. Temperature $> 0$ may produce more variable confidence distributions.

\textbf{Model scale.} 3--9B parameters only. \citet{kumaran2026} reported graded confidence distributions in Gemma 3 27B.

\textbf{Substrate.} TriviaQA factual QA only.

\textbf{Quantisation.} All models at Q5\_K\_M. \citet{cacioli2026g} showed AUROC$_2$ is robust to quantisation.

\textbf{Categorical prompt.} The accuracy collapse may be specific to the 10-class format, its anchor wordings, or the ``NO REASONING'' instruction.

\section{Pre-registration and data availability}

Pre-registered on OSF (registration: \url{https://osf.io/azbvx}, locked 15 April 2026; project: \url{https://osf.io/xgt73}). All data, code, and pre-registration document available on OSF and GitHub (\url{https://github.com/synthiumjp/koriat}). One deviation is disclosed (\S2.6). No hypotheses, thresholds, or decision rules were modified post-registration.

\appendix
\section{Validity Screening Protocol}

This appendix provides a self-contained specification of the psychometric validity screening protocol. The protocol was developed in \citet{cacioli2026d}, specified as a portable procedure in \citet{cacioli2026e}, and criterion-validated in \citet{cacioli2026f}.

\subsection{Input and binarisation}

The protocol operates on trials with a correctness label and a confidence value. Continuous confidence is binarised at 0.50, producing a 2$\times$2 contingency table:

\begin{center}
\begin{tabular}{lcc}
\toprule
& Correct & Incorrect \\
\midrule
High confidence ($\geq 0.50$) & $a$ & $b$ \\
Low confidence ($< 0.50$) & $c$ & $d$ \\
\bottomrule
\end{tabular}
\end{center}

\subsection{Degeneracy pre-check}

If the confidence signal has fewer than 3 distinct values or more than 95\% of binarised responses fall in a single category, the signal is classified Invalid (degenerate) without further analysis.

\subsection{Validity indices}

$L = b / (b + d) = P(\text{high confidence} \mid \text{incorrect})$. $L \geq 0.95$: at most 5\% error-detection rate.

$F_p = c / (a + c) = P(\text{low confidence} \mid \text{correct})$. $F_p \geq 0.50$: majority of correct trials receive low confidence.

$\text{RBS} = F_p - (1 - L)$. Positive values indicate directional inversion.

$\text{TRIN} = \max(n_{\text{high}}, n_{\text{low}}) / N$. Structural indicator; does not trigger Invalid alone.

$r(\text{confidence}, \text{correct})$ = point-biserial correlation. Diagnostic statistic; no classification action.

\subsection{Ordered screening sequence}

\begin{enumerate}
\item Degeneracy pre-check. If triggered, Invalid.
\item Cell counts. If any cell $< 5$, Insufficient.
\item TRIN. Report value. If $\geq 0.95$, structural warning.
\item $F_p$. If $\geq 0.50$ with Wilson CI lower bound $> 0.40$, Invalid.
\item $L$. If $\geq 0.95$ with Wilson CI lower bound $> 0.90$, Invalid.
\item RBS. If $> 0$ with CI excluding zero and point $> 0.05$, Invalid.
\item $r(\text{confidence}, \text{correct})$. Report value, $p$, 95\% CI.
\end{enumerate}

\subsection{Three-tier classification}

\textbf{Invalid.} Clear threshold violation. Signal uninformative.

\textbf{Indeterminate.} Threshold violated but CI includes valid values. Interpret with caution.

\textbf{Valid.} No threshold violations. Proceed to substantive analysis.

\bibliographystyle{plainnat}

\end{document}